\documentclass[conference]{IEEEtran}
\IEEEoverridecommandlockouts
\usepackage{cite}
\usepackage{amsmath,amssymb,amsfonts}
\usepackage{algorithmic}
\usepackage{graphicx}
\usepackage{textcomp}
\usepackage{xcolor}

\usepackage{caption}
\usepackage{subcaption}  

\usepackage{booktabs}
\usepackage{makecell}
\usepackage{soul}

\def\BibTeX{{\rm B\kern-.05em{\sc i\kern-.025em b}\kern-.08em
    T\kern-.1667em\lower.7ex\hbox{E}\kern-.125emX}}

\begin{document}

\title{Multi-Agent Reinforcement Learning for Dynamic Ocean Monitoring by a Swarm of Buoys\
\thanks{This research is partially funded by the SUTD SGP-AI Grant \#SGPAIRS1813}
}
%
\author{\IEEEauthorblockN{Maryam Kouzehgar}
\IEEEauthorblockA{\footnotesize \textit{Singapore University of Technology and Design}\\ 
Singapore \\
maryam\_kouzehgar@sutd.edu.sg}
\and
\IEEEauthorblockN{Malika Meghjani}
\IEEEauthorblockA{\footnotesize \textit{Singapore University of Technology and Design}\\ 
Singapore \\
malika\_meghjani@sutd.edu.sg}
\and
\IEEEauthorblockN{Roland Bouffanais}
\IEEEauthorblockA{\footnotesize \textit{Singapore University of Technology and Design} \\
Singapore \\
bouffanais@sutd.edu.sg}
}
\maketitle

\begin{abstract}
Autonomous marine environmental monitoring problem traditionally encompasses an area coverage problem which can only be effectively carried out by a multi-robot system. In this paper, we focus on robotic swarms that are typically operated and controlled by means of simple swarming behaviors obtained from a subtle, yet ad hoc combination of bio-inspired strategies. We propose a novel and structured approach for area coverage using multi-agent reinforcement learning (MARL) which effectively deals with the non-stationarity of environmental features. Specifically, we propose two dynamic area coverage approaches: (1) swarm-based MARL, and (2) coverage-range-based MARL. The former is trained using the multi-agent deep deterministic policy gradient (MADDPG) approach whereas, a modified version of MADDPG is introduced for the latter with a reward function that intrinsically leads to a collective behavior. Both methods are tested and validated with different geometric shaped regions with equal surface area (square vs. rectangle) yielding acceptable area coverage, and benefiting from the structured learning in non-stationary environments. Both approaches are advantageous compared to a na\"{i}ve swarming method. However, coverage-range-based MARL outperforms the swarm-based MARL with stronger convergence features in learning criteria and higher spreading of agents for area coverage. 
\end{abstract}


\section{Introduction}
\label{sec:intro}
Robotic systems are increasingly being used to perform essential data-gathering tasks by scientists~\cite{Matt2012}. For instance, in oceanography, robots and adaptive sensor networks are deployed with the aim of achieving a range of critical missions, such as monitoring climate variables, exploring the depth of ocean, keeping track of harmful algae blooms or other pollutant spills~\cite{song2015}. 


Traditionally, oceanographic missions have been accomplished using either moored buoys or fixed networks of bulky, partially submerged platforms. Such setups are costly both in terms of installation and maintenance. They also suffer from serious limitations in terms of spatial and temporal resolutions of the sensed data, especially when considering problems of permanent monitoring of large water bodies~\cite{zoss2018}. To overcome these critical constraints and limitations, the robotic community has considered a new paradigm focusing on developing simpler, smaller, low-cost, motorized, autonomous surface vehicles~\cite{yang2018grand}. A large team of such robotic units can collectively and dynamically be deployed at considerably lower cost and operational complexity than traditional monolithic systems~\cite{bouffanais2016design}. Such swarming systems---developed with a truly decentralized, scalable and fault-tolerant multi-robot system---are known to exhibit unprecedented effectiveness in operating in unknown, dynamic and unstructured environments, such as oceans. However, this transition from a robot-centric design to a system-centric one requires to consider critical elements beyond the electro-mechanical aspects at the robotic unit level~\cite{kit2019decentralized}. It can be argued that the most critical design component is the set of behavioral rules leading to more or less effective emergent collective actions by the system, i.e. swarm intelligence.

In previous works\cite{chamanbaz2017,zoss2018}, we presented the design, construction, and testing of a swarm of identical autonomous buoys, as well as a heterogeneous swarm~\cite{franc-oc2018} for dynamic monitoring operations. In \cite{chamanbaz2017}, we reported a range of classical swarming behaviors---aggregation, geofencing, heading consensus---while in \cite{zoss2018}, we covered adaptive environmental monitoring in dynamic regions. As argued earlier, one key system-level design aspect, in terms of their effectiveness in performing a monitoring task in a fully autonomous and distributed fashion, depends on our ability to identify an appropriate set of behavioral rules suited for a particular cooperative control strategy. 

In other previous works, we reported advances in speeding up reinforcement learning strategies using experience replay~\cite{karimpanal2018experience} and self-organizing maps~\cite{karimpanal2019self} for single-robot operations. In this work, we propose a novel and systematic approach towards that goal, which is based on applying multi-agent reinforcement learning (MARL) to identify a set of behavioral rules that are generating dynamic and responsive collective behaviors while dealing with partial observability and non-stationarity of environmental features.

Furthermore, area coverage has broad applications in oceanography, especially for monitoring, surveillance and search and rescue applications \cite{meghjani2016}. If the swarm robotic system is used to gather environmental data and features such as temperature and salinity, the agents should be able to spread as uniformly as possible across the region of interest, i.e. performing a ``blanket coverage'' or dynamic area coverage. In contrast, if the system is tasked to track how a substance spreads---e.g. oil spills and harmful algae blooms---it should be endowed with the ability to autonomously redeploy itself toward the border of the region of interest, i.e. achieving dynamic geofencing.
The blanket coverage scenario for dynamically changing surface areas has been considered in~\cite{zoss2018} based on swarming rules obtained from a subtle, yet totally ad-hoc, combination of ``avoidance" and ``attraction" as defined in the widely used Alignment--Attraction--Avoidance (AAA) model for biological swarms. 

\begin{figure*}[t!]
    \centering
    \includegraphics[width=\textwidth]{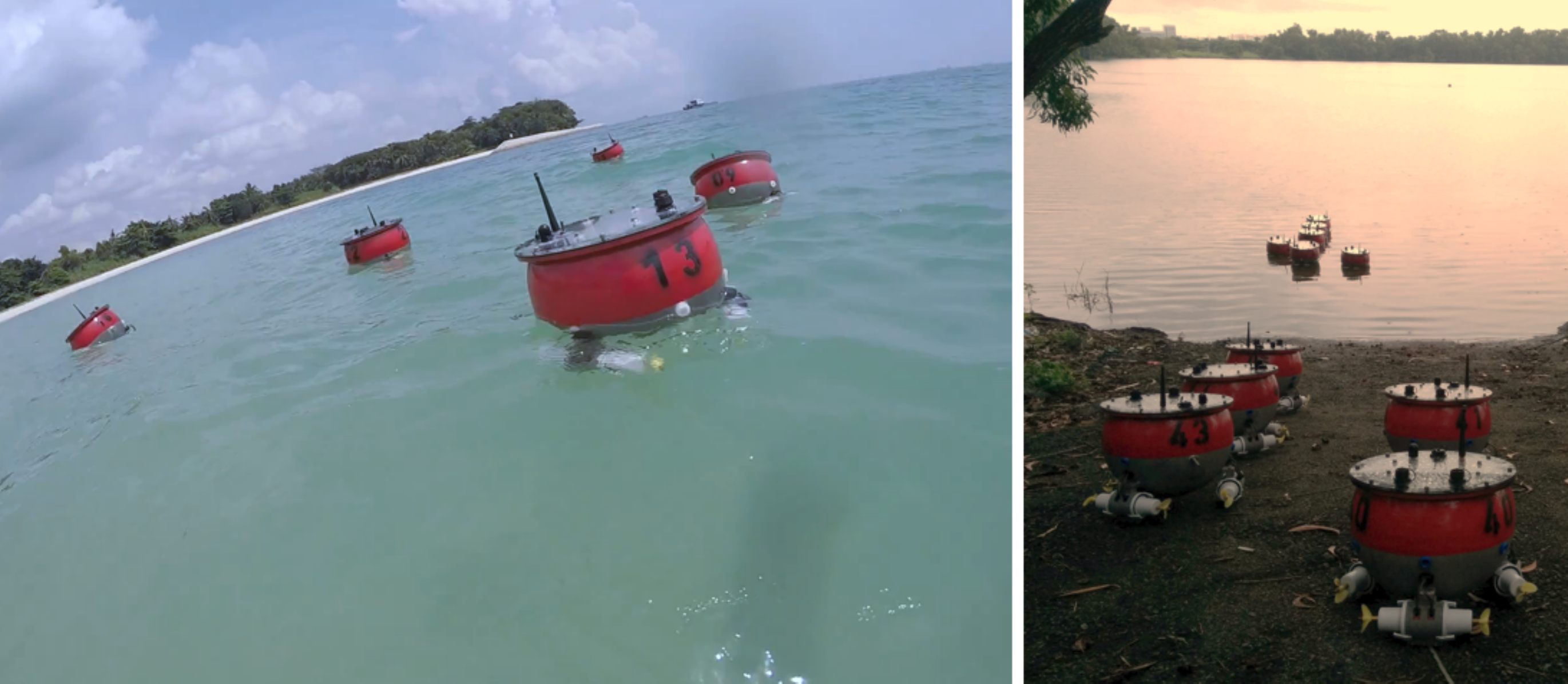}
    \caption{The BoB system deployed during field tests in a freshwater reservoir in Singapore.}
    \label{fig1}
\end{figure*}

Beyond these ad-hoc ways of designing swarm behavioral rules~\cite{chamanbaz2017,franc-oc2018,zoss2018}, here we consider MARL as a way to learn collectively more effective group-level responses for a swarm of agents. The simulations presented in this study can readily be implemented on our swarm of buoys referred to as “Bunch of Buoys” (or BoB for short) when performing area coverage of surface areas with different shapes. With our particular swarm of buoys (see Fig.~\ref{fig1}), the individual units are forming an ad-hoc communication network enabling the seamless addition or removal of any number of nodes while the system is operating. This critical feature of distributed mesh communication is achieved using XBee-PRO modules as part of what we call the swarm-enabling unit~\cite{chamanbaz2017}. Essentially, the network topology is time-varying and it automatically reconfigures itself as the agents move and enter or leave each other’s neighborhood, i.e. communication range.
The original buoy (BoB or BoB-0 in~\cite{franc-oc2018}) features a compact, omnidirectional, watertight design~\cite{zoss2018}. A vectored propulsion system implemented with three pairs of motors allow the buoy to move through water surfaces at speeds of up to 1.0 m/s. The buoy is capable of self-localization using a GPS module and is designed to host a range of sensors to characterize its local environment.


In the recent literature dealing with MARL, two distinct research directions can be found: A majority of the works consider fully centralized reinforcement learning among multiple agents~\cite{Gupta2017}, some are concerned with centralized learning and decentralized execution~\cite{maxim2018, maxim2019, Lowe2017, Forester2018}, and very few references are focusing on fully decentralized approaches~\cite{Zhang2018}. In addition, there are also works on merging concepts from centralized and decentralised MARL at the level of underlying utilized networks such as ~\cite{Forester2018} that uses a centralised critic for estimating the $Q$-function and decentralised actors for policy optimization. Further studies and attempts using Actor-Critic networks are reported in the literature~\cite{Sutton1998, Lowe2017, Mnih2016asyn}, which are in direct relation with the considered MARL algorithm in this paper.      

In this study, we are primarily interested in performing a particular application---namely area coverage for ocean monitoring, and we will be using MARL algorithms as provided with slight adaptations and modifications. Our contribution lies in the marine applications to swarming systems rather than manipulating the training algorithms. We will be defining the state-action environment and propose promising reward functions for dynamic area coverage. The utilized MARL framework follows the Multi-Agent Deep Deterministic Policy Gradients (MADDPG) provided in~\cite{Lowe2017}. In order to put forward a MARL scenario, we define a position-based state-space relying on the fact that the buoys are capable of self-localization using their GPS module. In addition, the action-space is defined based on robot movements, similar to the works presented in~\cite{malika2016, Lowe2017, Gupta2017, openai2017}. Finally, the rules required for dynamic area coverage will be embedded in the design of reward function for the collective agents. We believe, the obtained solution with the learning agents in the swarm is promising for dynamic and adaptive area coverage.

The rest of the paper is organized as follows. Section \ref{sec:MADDPG} briefly elaborates on the MADDPG algorithm. Section \ref{sec:PropApp} is concerned with the details of the proposed approach casting light on state-space, action-space and design of reward functions for the proposed MARL-based area coverage methods. This section also elaborates on our modification of the MADDPG algorithm. Section \ref{sec:results and discussion} is dedicated to simulation results with the proposed approach and also presents a simulation on a real waterbody located in Singapore: Bedok Resrvoir. Lastly, Section \ref{conc} concludes the paper and provides potential future research directions.

\section{MADDPG Algorithm: A brief review}
\label{sec:MADDPG} 
Non-stationarity of the environment is one of the main challenges in MARL problems, because as training progresses, each agent’s policy is changing and thus the environment becomes non-stationary from the perspective of any individual agent.

The MADDPG algorithm was proposed as an extension of actor-critic policy gradient methods (DDPG)~\cite{Lili2015} in which the critic is augmented with extra information about the policies of other agents, while the actor only has access to local information~\cite{Lowe2017}. Utilizing this augmented critic, or fully observable critic is a common solution to address non-stationarity of the environment. Thus, this centralized critic can be used as a trusted guide for local actors to improve flexibility. In contrast, generally, MADDPG follows a promising Centralized Training, Decentralized Execution (CTDE) approach i.e. after training is completed, (with centralized critic) only the local actors are used at execution phase. 
\subsection{Deep Deterministic Policy Gradient method (DDPG)}
With MADDPG, each agent is trained by a DDPG algorithm in which the actor has access to the local observations while the centralized critic $Q$ concatenates all states-actions together as the input and uses the local reward to obtain the corresponding $Q$-value. The critic is trained by minimizing a DQN-like loss function.

It is worth reminding that, the main idea in the Policy Gradient approach (PG) is to directly adjust the parameter $\theta$ of the policy $\pi$ to maximize the objective $J(\theta)$ given in \eqref{eq:J} by taking steps in the direction of gradient given by \eqref{eq:grad} \cite{Sutton2000}. The gradient is calculated based on ``Policy Gradient Theorem", whose proof is provided in~\cite{Sutton1998}. The notations for the symbols used in the following equations are summarized in Table.~\ref{tab1}. 
\begin{equation}\label{eq:J}
\textcolor{white}{\mathbb{E}}
J(\theta) =\mathbb{E}_{s\thicksim p^{\pi},a\thicksim \pi_{\theta}}[\mathcal{R}],
\end{equation}
\begin{equation}\label{eq:grad}
\nabla_{\theta} J(\theta) =\mathbb{E}_{s\thicksim p^{\pi},a\thicksim \pi_{\theta}} [\nabla_{\theta}\log \pi_\theta(a|s)Q^{\pi}(s,a)],
\end{equation}
It is also possible to extend the policy gradient framework to deterministic policies $\mu_\theta: S \rightarrow A$. Since this theorem relies on $\nabla Q_\mu(s, a)$, it requires that the action space $A$ (and thus the policy $\mu$) be continuous. In particular, under certain conditions we can write the gradient of the objective as follows:
\begin{equation}\label{eq:Dgrad}
\nabla_{\theta} J(\theta) = \mathbb{E}_{s\thicksim \mathcal{D}} [\nabla_{\theta}\mu_\theta(a|s)\nabla_{a}Q^{\mu}(s,a)|_{a=\mu_{\theta}(s)}].
\end{equation}
Finally, DDPG (Deep deterministic policy gradient) is a variant of DPG where the policy $\mu$ and critic $Q^\mu$ are approximated with deep neural networks: Actor network receives observation (state variables) and citric value at the input layer, and after processing it through several hidden layers, outputs the action; the critic network receives the observation and action, processes it through hidden layers and ultimately outputs the $Q$ value.
\begin{table}[bp]
    \centering
    \caption{\small \\
    \textsc{Used Notations}}
        \begin{tabular}{|c c|} 
        \toprule
		\textbf{\textit{Symbol}} & \textbf{\textit{Definition}} \\ 
        \midrule
        \textbf{$s\in \mathcal{S}$}  & States \\
        \textbf{$a\in \mathcal{A}$}  & Actions \\
        \textbf{$r\in \mathcal{R}$}  & Rewards \\
        \textbf{$\gamma$} & \makecell{Discount factor: penalty to uncertainty\\ of future rewards $0 <\gamma \leq 1$.} \\
        \textbf{$\eta$} & learning rate $0 <\eta \leq 1$. \\
        \textbf{$\pi(a|s)$}  & Stochastic policy (agent behavior strategy) \\
        \textbf{$\pi_\theta(.)$} & a policy parameterized by $\theta$ \\
        \textbf{$a=\mu(s)$} & $\mu$ is a deterministic policy: $S \rightarrow A$ \\ 
        \textbf{$Q(s,a)$} &	\makecell{Critic or Action-value function\\ assessing the expected return of\\ a pair of state and action (s,a)}\\
        
        \hline
        \end{tabular}
    \label{tab1}
\end{table}

\subsection{Original Multi-Agent DDPG (MADDPG)}
\begin{figure}[tp]
\centerline{\includegraphics[width=85mm]{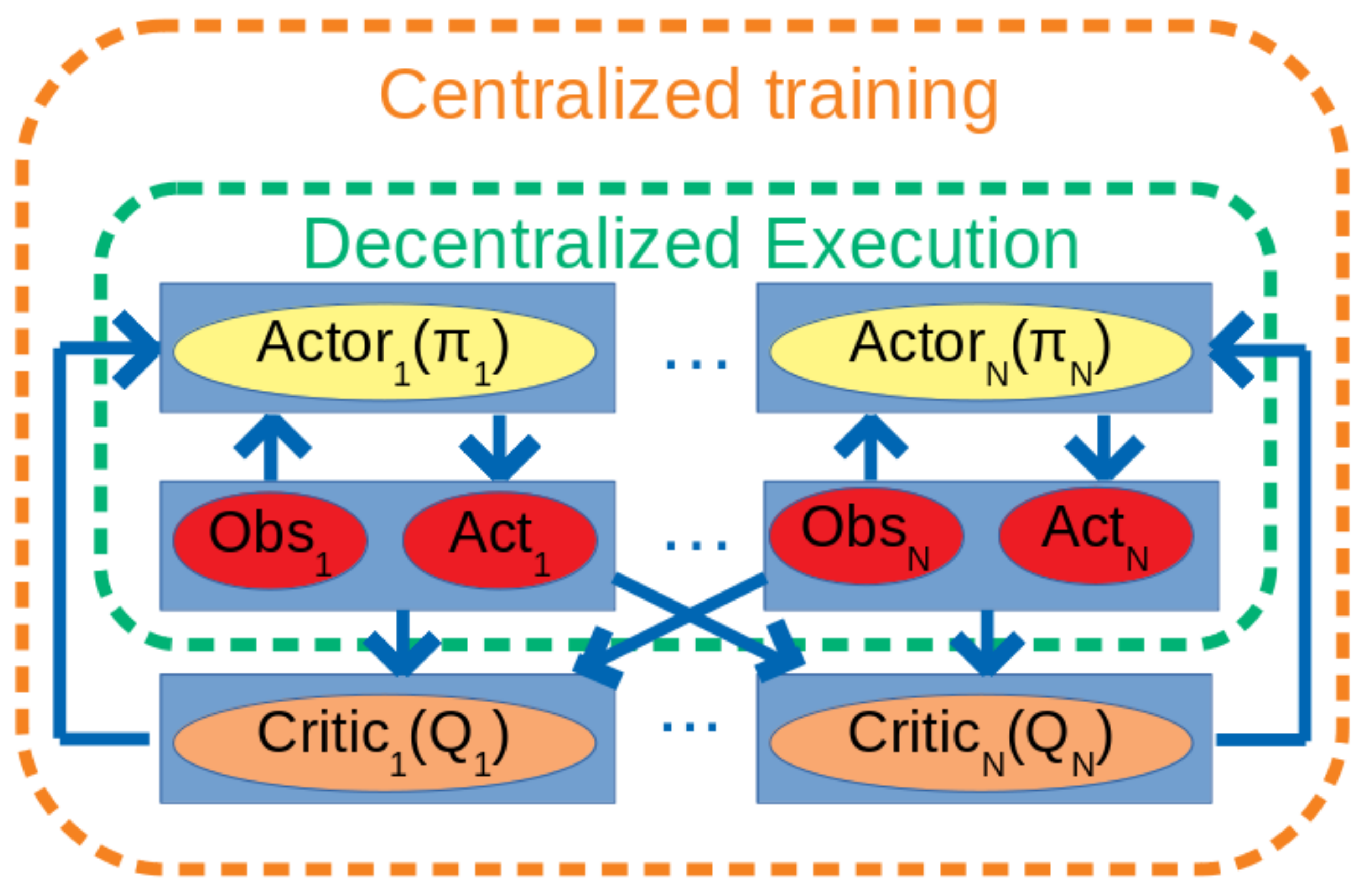}}
\caption{A  schematic diagram of the MADDPG algorithm}
\label{MADDPG-schematics}
\end{figure}

For application to multi-agent dynamics, a schematic diagram of the original MADDPG algorithm is shown in Fig.~\ref{MADDPG-schematics} where it is also emphasizing on CTDE style. 

As it is obvious, MADDPG is following an actor-critic style in which the training is considered in a continuous observation and action space. While training the actor and critic networks, the algorithm takes into account joint observations and actions of all agents, and at the time of execution, the actor network of each agent takes \textit{only local} observation into account. 
The joint observations of all agents is denoted by $\mathbf{x} = (o_{1}, ..., o_{N})$, and joint actions is $a = (a_{1}, ..., a_{N})$. The joint observations and actions are stored in a replay buffer $\mathcal{D}$ together with ``shared reward" $r$, 
in the form of $(\mathbf{x}, a, r, \acute{\mathbf{x}})$ where $\acute{\mathbf{x}}$ is the new state (observation). A random mini batch of S
samples $(\mathbf{x}^{j}, a^{j}, r^{j}, \acute{\mathbf{x}}^{j})$ is extracted from $\mathcal{D}$, and based on \eqref{eq:y} the critic network for each agent is updated by minimizing the loss function given in \eqref{eq:loss}. Furthermore, the actor network for each agent, is updated by a sampled policy gradient given in \eqref{eq:actor-update}.
Finally, similar to DDPG, the target networks are soft updated at every certain steps by $\acute{\theta_{i}} \leftarrow \tau \theta_{i}+(1-\eta)\acute{\theta_{i}} $~\cite{Lowe2017}.
\begin{align}
y^{j} &= r^{j}_{i}+\gamma Q_{i}^{\acute{\mu}}(\acute{\mathbf{x}}^{j},\acute{a}_{1},..., ,\acute{a}_{N})\mid_{\acute{a}=\acute{\mu}(o^{j})},\label{eq:y}\\
\mathcal{L}(\theta_{i})&=\frac{1}{\mathcal{S}}\sum_{j}\left(y^{i}- Q_{i}^{\mu}(\mathbf{x}^{j},a_{1}^{j},...,a_{N}^{j})\right)^2,\label{eq:loss}\\
\nabla_{\theta_{i}} J &\approx \frac{1}{\mathcal{S}}\sum_{j} \nabla_{\theta_{i}}\mu_{i}(o_{i}^{j})\nabla_{a_{i}}Q_{i}^{\mu}(\mathbf{x}^{j},a_{1}^{j},...,a_{N}^{j})\mid_{a_{i}=\mu_{i}(o_{i}^{j})}, \label{eq:actor-update}
\end{align}

\section{Proposed Approach}\label{sec:PropApp}
In this paper, a modified version of the MADDPG algorithm~\cite{Lowe2017} is proposed considering the Multi-Agent Particle Environment introduced in~\cite{mpe2017} in which a continuous observation and action space is considered along with some basic simulated physics to control movements in the environment. A total of $9$ agents is considered to cover a square region of edge length $3$~m and a rectangular region of 1~m$\times$9~m. Agents are released in a confined corner of the environment with random positions and throughout MARL they learn to spread so as to achieve coverage of the region.  

In simulations, the state space is considered as a 2D space for agent positions, i.e. the observations are 2D vectors of position, and the action-space is the continuous movement commands in 2D directions, i.e. difference between current position and destination position in both dimensions.

After defining the state-space and action-space based on \cite{mpe2017}, and in order to fulfill the area coverage scenario for oceanography, we propose two types of reward functions inspired by recent works on swarming behaviors~\cite{zoss2018} and deep MARL~\cite{maxim2018}. The definition of the reward functions, the simulation results, and performance analysis are given below. 
\subsection{MARL using swarming (SW-MARL)}
As mentioned above, MARL is as a systematic approach based on learning and is capable of handling unknown environments and non-stationarity and thus benefits from several advantages in comparison to swarming which is based on ad-hoc ways of designing swarm behavioral strategies. In order to implement SW-MARL, we consider the widely-used combination of ``attraction" and ``avoidance" as defined in the AAA framework for biological swarms~\cite{zoss2018} in a fixed pre-defined region of interest as required in most oceanographic studies.

\paragraph{Attraction toward inside the region} Inspired by \cite{zoss2018}, for ``attraction", a positive reward is considered for the agents to attract them towards the interior of the area as expressed in \eqref{eq:sw-RegAtt} with $R_{1}=150$, $R_{2}=200$ where $D(x,y) < 0$ is considered as the definition of the interior of the desired region and a distance-based negative reward is considered as agents move away from one another. In other words, $D$ is a signed distance function or at least a function that increases monotonically outside the region. Geometrically, it can be interpreted by the mathematical definition of borders of region for example for a disk region with radius $r$ centered at $(0,0)$, we have $D(x_{a},y_{a}) = x_a^2 + y_a^2 -r^2$.
\begin{equation}\label{eq:sw-RegAtt}
    \text{Reward}(a) =
    \begin{cases}
        +R_{1} & \text{if $D(x,y) < 0$}, \\
        - R_{2}-D(x,y) & \text{else.}
    \end{cases}
\end{equation}

\paragraph{Inter-agent Attraction/Avoidance} Inspired by \cite{maxim2018}, we define an \textit{inter-agent collision (avoidance)} range and based on this range, a negative reward is introduced when any two agents are too near to one another, and on the other hand, we define \textit{inter-agent proper} range, and based on this range, a positive reward is considered for being in certain acceptable range from each other. These acceptable ranges are basically defined based on agent size and can be generalized to encompass concepts such as sensory range. Thus based on \eqref{eq:sw-InterAg}, each agent gets a positive reward for being in proper distance to others and a negative reward for being in too close contact with others. In \eqref{eq:sw-InterAg}, $N$ is the number of agents and $d(n,i)$ denotes the distance between agent $n$ and agent $i$.

\begin{equation}\label{eq:sw-InterAg}
    \begin{split}
    \text{Inter\_Agent\_Reward} = \\
    & \alpha \sum_{i=1}^{N}\sum_{n>i}^{N} \text{in\_prop\_range}(d(n,i))\\ 
    & -\beta\sum_{i=1}^{N}\sum_{n>i}^{N} \text{in\_coll\_range}(d(n,i)),
    \end{split}
\end{equation}

in which we have:
\begin{equation}\label{eq:3}
  \text{in\_prop\_range}_{[a,b]}(x) = 
   \begin{cases}
        1 & \text{if $x \in [a,b] $}, \\
        0 & \text{else.}
    \end{cases}
\end{equation}

\begin{equation}\label{eq:4}
  \text{in\_coll\_range}_{[c,d]}(x) = 
   \begin{cases}
        1 & \text{if $x \in [c,d] $}, \\
        0 & \text{else}.
    \end{cases}
\end{equation}
where we choose $\alpha = 150, \beta = 200$, $[a,b]=[0.85,1.2]$ and $[c,d]=[0, 0.56]$ with agent radius $r=0.08$. The values of the coefficients of the reward function are empirically selected and tuned based on a pool of simulation results.

\paragraph{Alignment} In the literature about swarms, alignment means the behavior of swarm while performing heading consensus and align themselves in a common direction of movement~\cite{zoss2018}. Since we seek to perform area coverage and the targeted region is static, in our case, alignment is not applicable, but in case we deal with a moving region, generalizing the current approach to implement alignment as well, would not be a challenge. 

For SW-MARL simulations, the original MADDPG algorithm is considered and this will be further illustrated in section~\ref{sec:results and discussion}.

\subsection{MARL using Coverage Range (CR-MARL) and modified MADDPG}
\subsubsection{CR-MARL}
To apply area coverage concepts in our current MARL framework, we define some environmental entities based on polygon concepts as follows:

\begin{itemize}
  \item \textit{Region polygon:} Any free-shape polygon can be defined to be considered as the target area to be covered.
  \item 
  \textit{Agent Coverage Polygon:} A fixed-radius circle-like polygon (a 30-edge polygon considered) around the agent which is basically representing the agent's sensory range for specific data collection tasks. In this study, this polygon is denoted by agent's ``coverage polygon" and its radius is labeled as agent ``coverage range".
\end{itemize}

\begin{figure}[bp]
\centerline{\includegraphics[width=75mm]{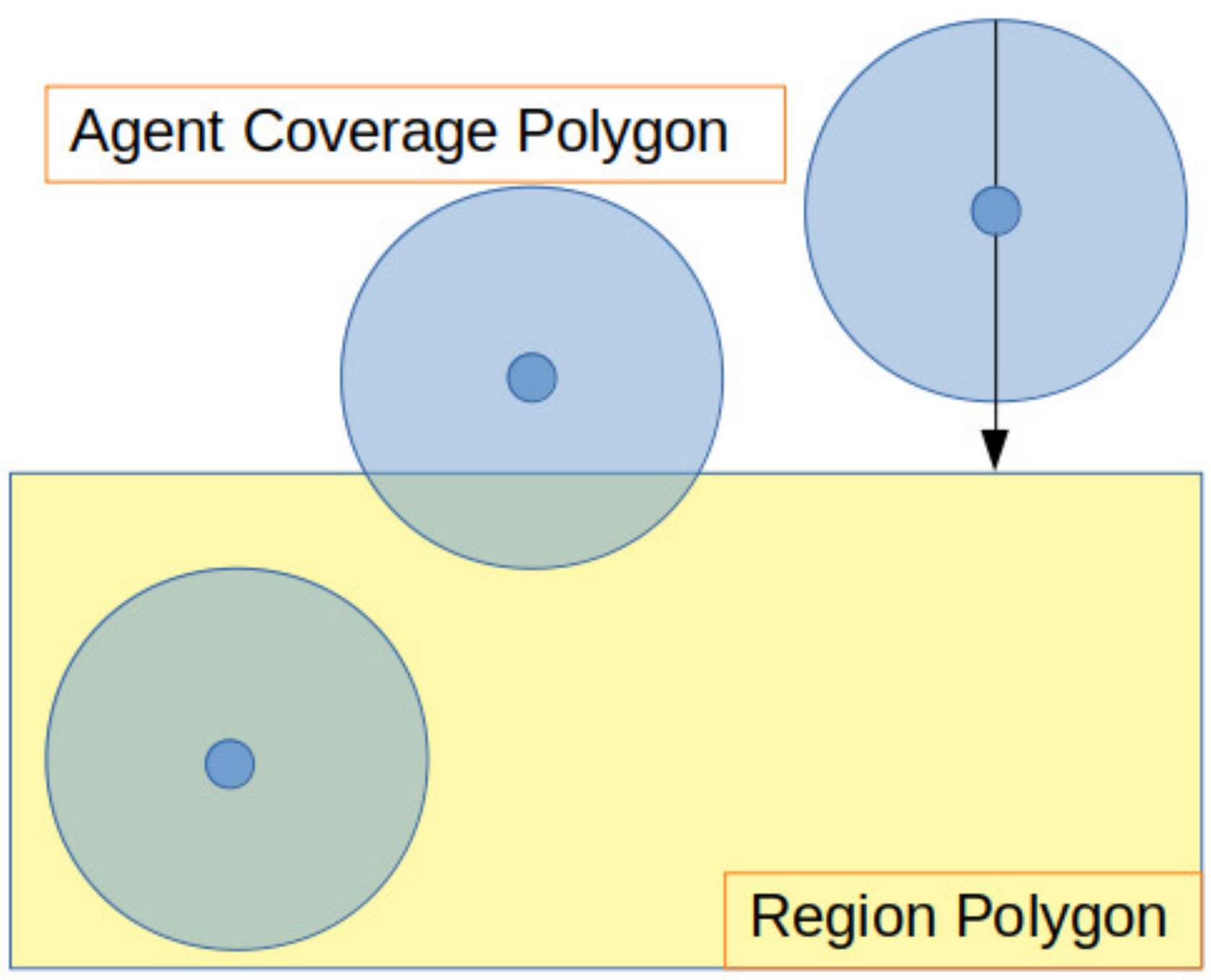}}
\caption{Three different positions of agent vs. region: (1) Falling inside, (2) Partially overlapping, and (3) Fully outside}
\label{pos}
\end{figure}

Furthermore, to insert these coverage concepts into the MARL reward function, three different positions of agent vs. region have been taken into account which are depicted in Fig.~\ref{pos} illustrating the two different types of coverage reward considered.

\paragraph{Agent-specific coverage  reward} Considering Fig.~\ref{pos}, an ``agent-specific coverage reward", is calculated based on \eqref{eq:5} where each agent is rewarded based on the amount of region it covers, and if it has no coverage (and does not exist inside region), there is a penalty proportional to the distance from the region. As we will show, this choice will fulfill the individual aspect of coverage reward. On the other hand, compared to SW-MARL, here  CR-MARL turns out to have a more efficient ``Attraction" toward the interior of the region as it benefits from stronger constraints attempting to keep not only the agent itself, but also the agent coverage polygon inside the region.

\textcolor{white}{.}

$\text{Cover\_Reward}(a) =$
\begin{equation}\label{eq:5}
            \begin{cases}
                +R_1 + A(\text{Cov}(a)) & \text{if Cov$(a) \subset $Reg}, \\
                +R_2 + A(\text{Intersect(Cov$(a)$,Reg)}) & \text{if Cov$(a) \cap$  Reg}\ne	\varnothing, \\
                -R_3-\text{dist(Reg,Cov(a))} & \text{if Cov$(a) \notin$ Reg.}
            \end{cases}
\end{equation}
where $A$ stands for Area, Cov($a$) shows ``Coverage Polygon" of agent $a$, Reg is the ''Region Polygon", agent radius is considered $r_{a}=0.08$~m, coverage polygon radius is assumed $R=0.5$~m and coefficients of reward function are defined as $R_1$ = 60~m$^2$, $R_2$ = 30~m$^2$, $R_3$ = 0~m supporting the logic that the more an agent's coverage polygon overlaps the region polygon, it gets more reward. The values of the coefficients of the reward function are empirically selected and tuned based on a pool of simulation results.

\begin{figure}[bp]
\centerline{\includegraphics[width=75mm]{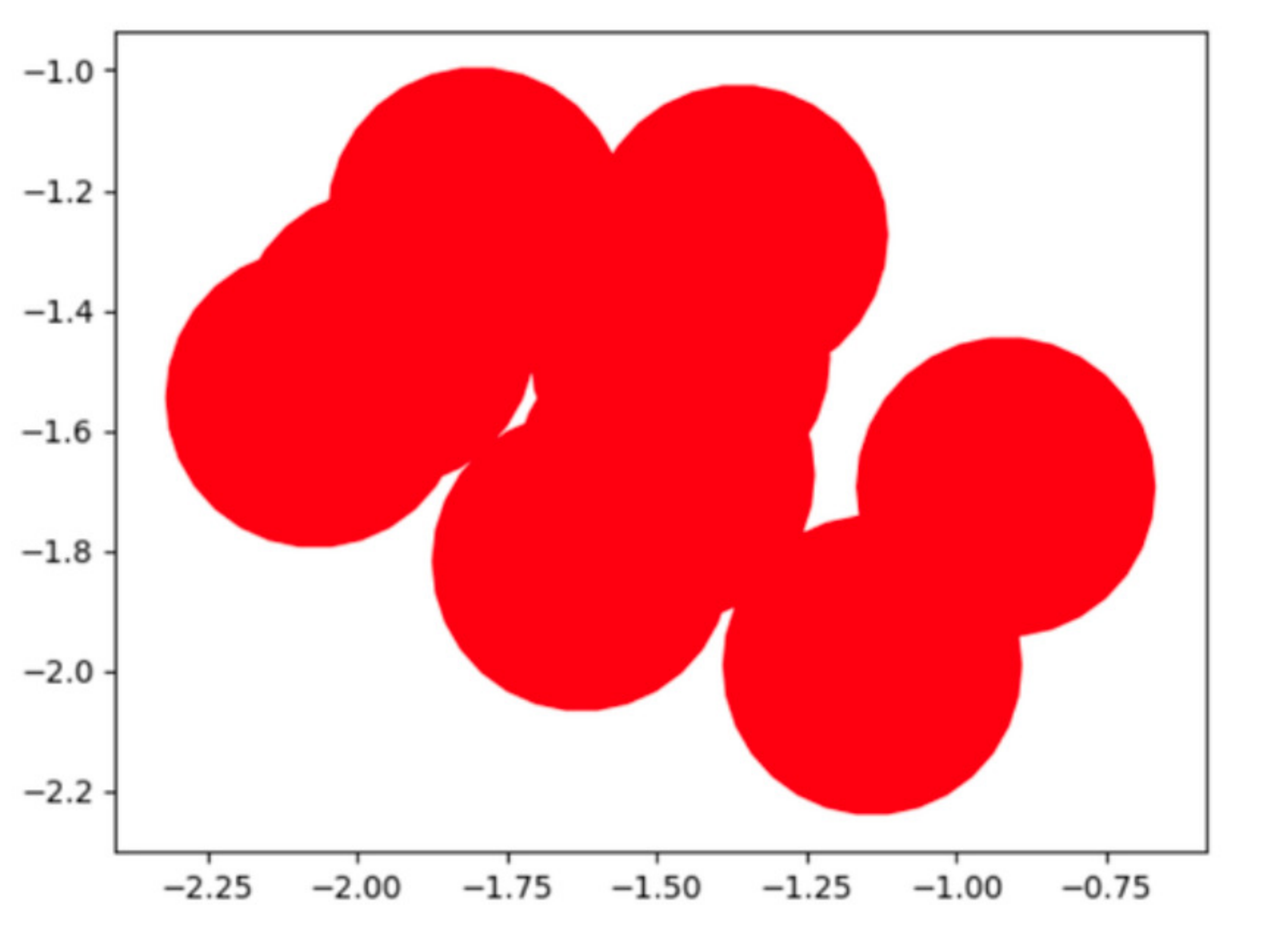}}
\caption{Overall coverage with merging overlaps representing collective reward}
\label{ovcov}
\end{figure}

\paragraph{Overall Coverage reward}
Furthermore, agents are positively rewarded based on the ``overall Area Coverage", regardless of the overlaps between coverage polygons. Overlaps are merged as shown in Fig.~\ref{ovcov}. This way, the collective aspect of reward is provided and the overall reward is the sum of collective and individual rewards as in \eqref{eq:6}, where the coefficients are considered as $c_1=1$ and $c_2=5$, thereby emphasizing the performance of a collective task.
\begin{equation}\label{eq:6}
\begin{split}
 & 
 \textcolor{white}{\Bigg( \Bigg. } 
    \text{Overall\_Reward(a)}=c_1*\text{Cover\_Reward}(a)\\ 
 & +c_2* \text{Overall\_Area\_Coverage.}
\textcolor{white}{\Bigg. \Bigg))} 
\end{split}
\end{equation}

\subsubsection{Modified MADDPG}
For the CR-MARL reward function introduced above, we have implemented a modified MADDPG in which the reward sharing policy is especially appropriate for the collective nature of the introduced reward function. In the original MADDPG, the agents share a collective reward $r$ (for example the sum of all agents’ rewards). In our approach, we have introduced the following modifications:
\begin{itemize}
    \item The proposed reward on its own, encompasses collective concepts and thus, the need for an explicit reward sharing is eliminated. Specifically, the reward in \eqref{eq:6} contains agent-specific reward and collective reward. The latter eliminates the need to use a shared reward, e.g. a summation of rewards of all agents.
    \item The structure of the collective reward depends on the arrangement of the whole swarm, and thus the overall reward by itself has a collective nature. Based on this, the modification of the original MADDPG is specifically applied to \eqref{eq:y}, in which, for the MADDPG collaborative environment $r_{i}$ is same for all the agents as a shared reward $r$ which can be implemented for example as a summation of rewards of all agents returned back to every agent.
    \item In our approach, agents share information via the intrinsic collective nature of the reward, and also they can sense their own share of the reward function. This is effectively acting collaboratively beyond blind reward sharing policy in original collaborative MADDPG.
    \item Furthermore, agents experience some levels of independence based on their individual reward.
\end{itemize}

\section{Simulation results and Discussion}
\label{sec:results and discussion}
For all the simulations, the considered regions have same area of $9~\text{m}^2$ and the MARL algorithm has been considered with 10,000 episodes.
\subsection{Results of the SW-MARL approach}
\begin{figure}[htbp]
\centering
    \begin{subfigure}[b]{0.99\linewidth}
        \centering
        \includegraphics[width=\linewidth]{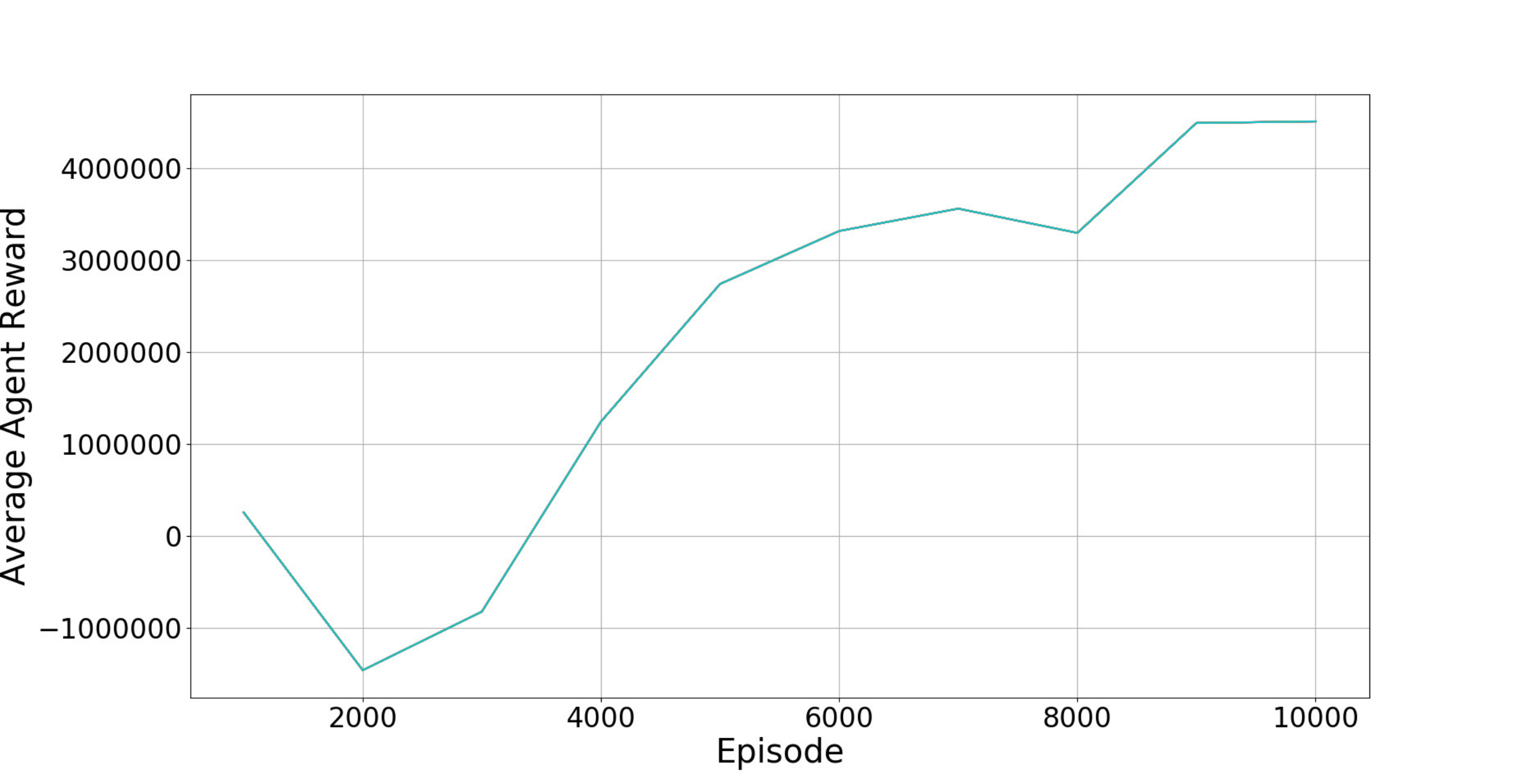}
        \caption{Average agent reward}
        \label{ave_swarmsq}
    \end{subfigure}
    \begin{subfigure}[b]{0.99\linewidth}
        \centering
        \includegraphics[width=0.45\linewidth]{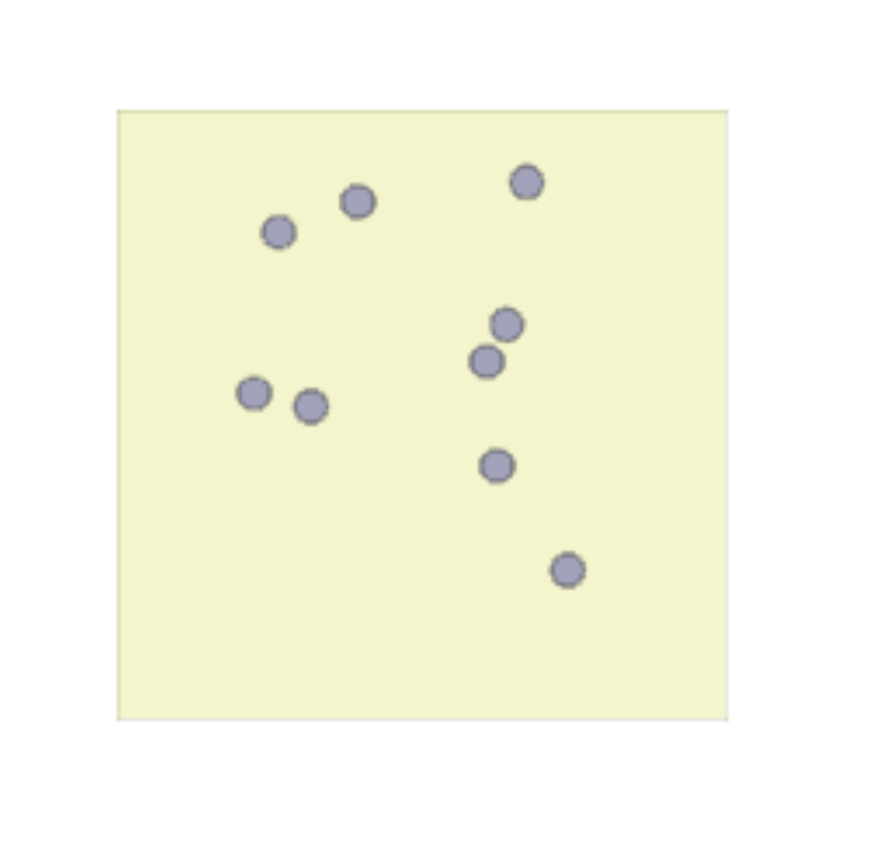}
        \caption{Qualitative performance for the area coverage}
        \label{swarmSQperform}
    \end{subfigure}
\caption{Results of SW-MARL for coverage of a square region.}
\label{swarm_sq}
\end{figure}

\begin{figure}[htbp]
\centering
    \begin{subfigure}[b]{0.99\linewidth}
        \centering
        \includegraphics[width=\linewidth]{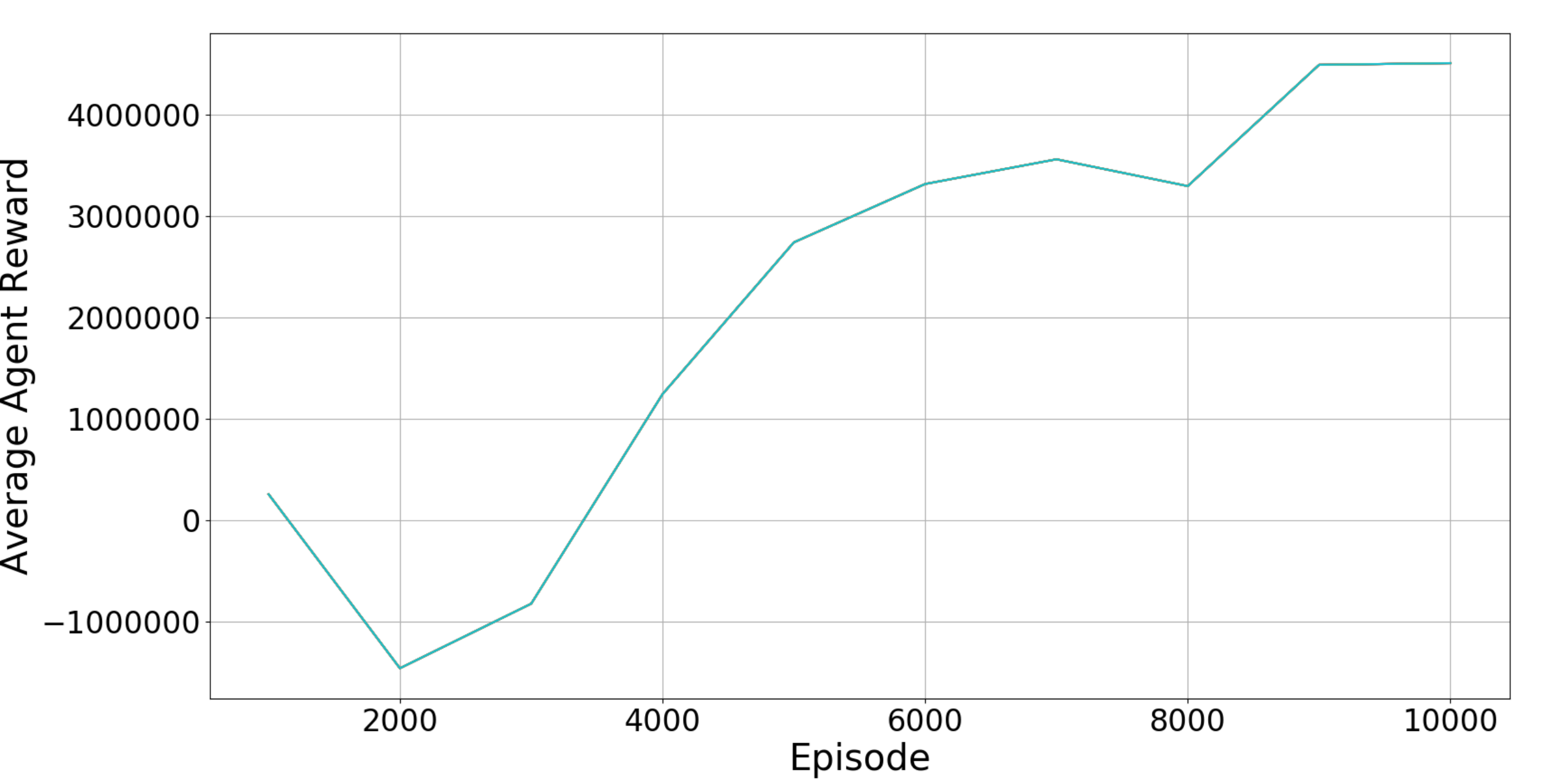}
        \caption{Average agent reward}
        \label{ave_swarmHorRect}
    \end{subfigure}
    \begin{subfigure}[b]{0.99\linewidth}
        \centering
        \includegraphics[width=\linewidth]{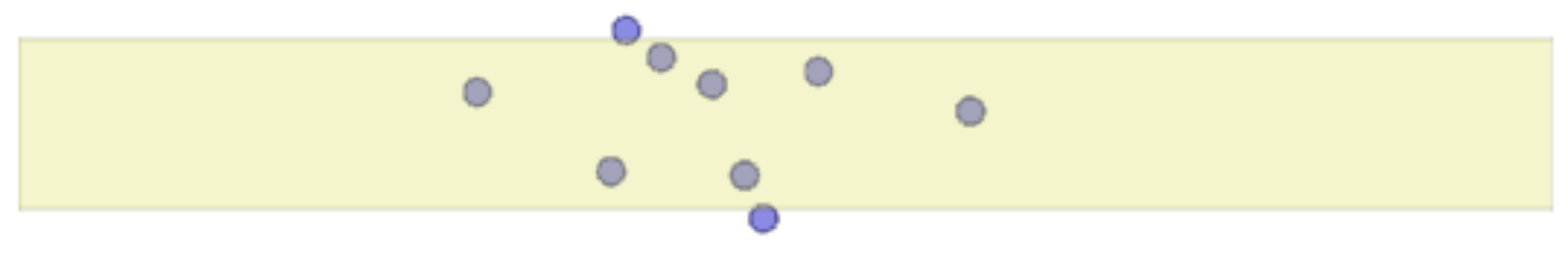}
        \caption{Qualitative performance for the area coverage}
        \label{sw_HorREct_perform}
    \end{subfigure}
\caption{Results of SW-MARL for coverage of a rectangle region.}
\label{swarm_HorREct}
\end{figure}

For this set of simulations with the swarming-based reward function, since the reward function lacks intrinsic collective concepts, the original MADDPG algorithm is considered with reward sharing for collaborative scenarios i.e. a summation of reward of all agents is returned back to every agent as the shared reward. The results of SW-MARL for square region and horizontal rectangle region are reported in Fig.~\ref{swarm_sq} and Fig.~\ref{swarm_HorREct} respectively. In Fig.~\ref{ave_swarmsq}, the average reward for each agent is shown throughout the training episodes, and Fig.~\ref{swarmSQperform} shows a result of the qualitative performance out of rendering environment. An identical explanation exists for all the following pairs of figures.
In Fig.~\ref{swarm_sq} and Fig.~\ref{swarm_HorREct}, since the SW-MARL is trained with original MADDPG, all agents share a collective reward which is shown by a single graph, thus the labeling of agents are not included.


\subsection{Results of CR-MARL approach}
\begin{figure}[htbp]
\centering
    \begin{subfigure}[b]{0.99\linewidth}
        \centering
        \includegraphics[width=\linewidth]{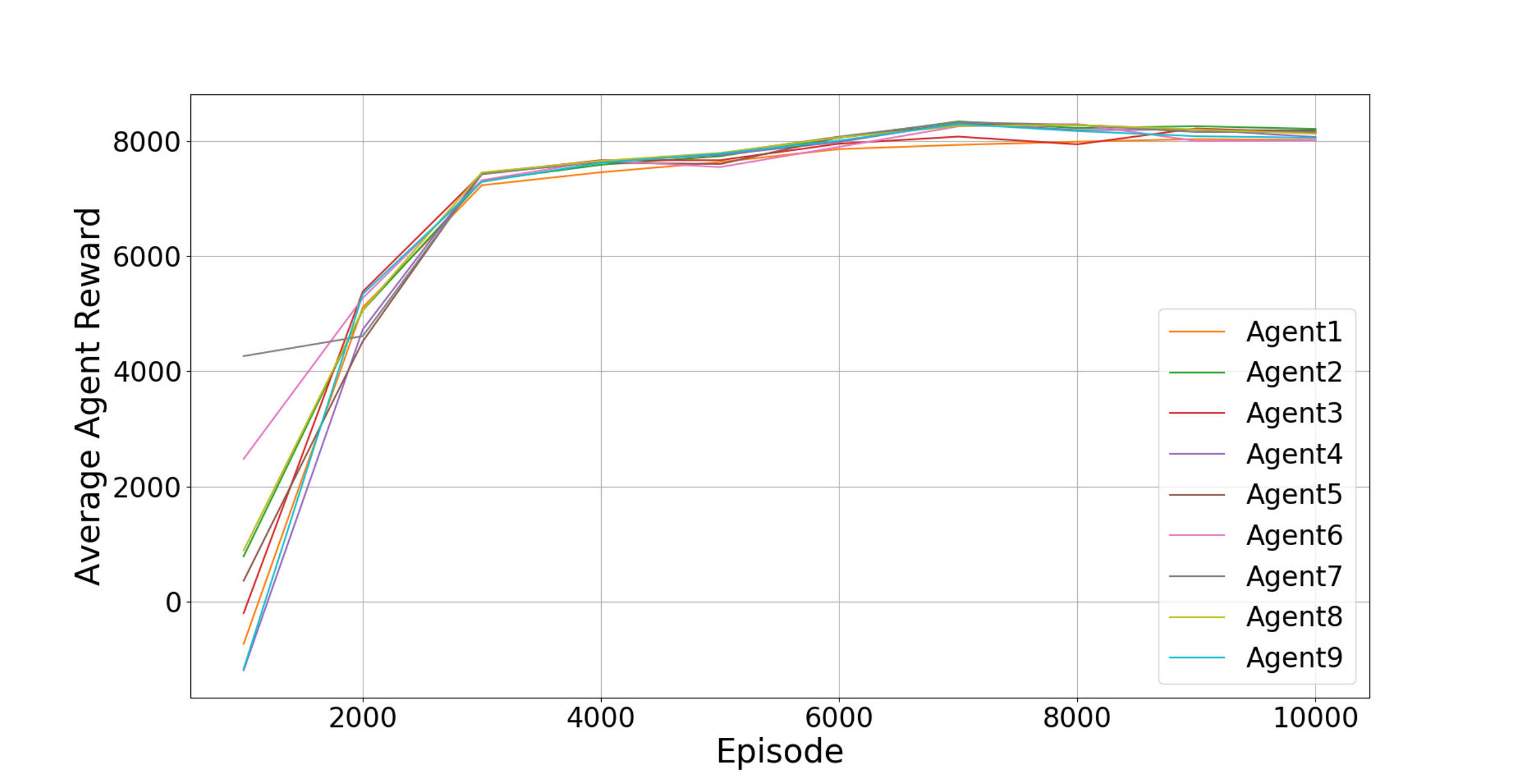}
        \caption{Average agent reward for each of the 9 agents constituting the multi-agent system}
        \label{ave_coverSq}
    \end{subfigure}
    \begin{subfigure}[b]{0.99\linewidth}
        \centering
        \includegraphics[width=0.46\linewidth]{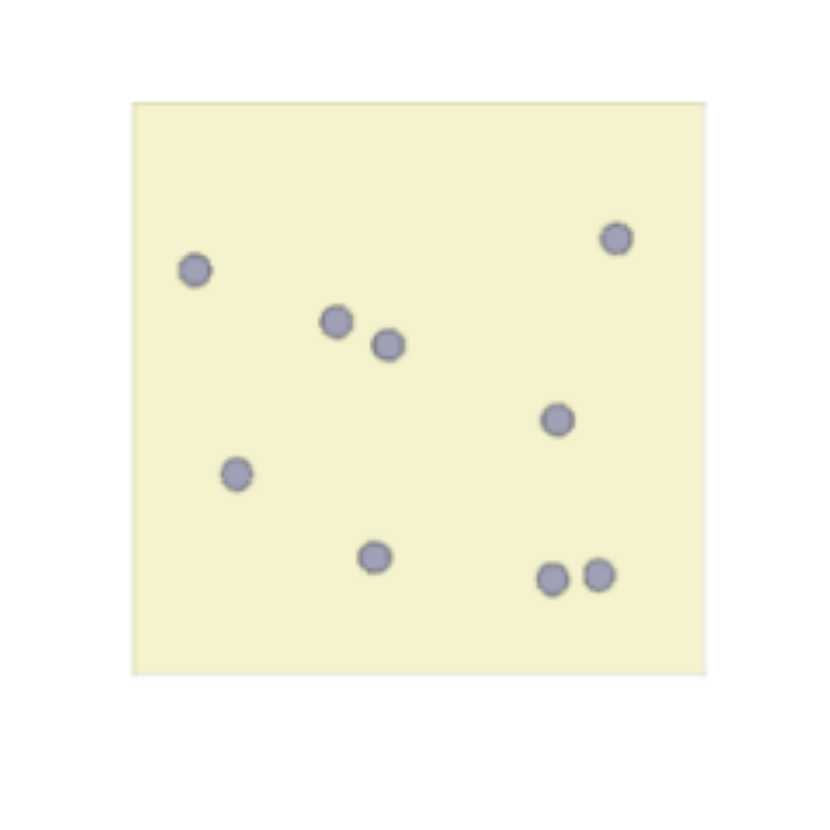}
        \caption{Qualitative performance for the area coverage}
        \label{cov_sq_perform}
    \end{subfigure}
\caption{Results of CR-MARL for coverage of a square region.} 
\label{cov_sq}
\end{figure}

\begin{figure}[htbp]
\centering
    \begin{subfigure}[b]{0.99\linewidth}
        \centering
        \includegraphics[width=\linewidth]{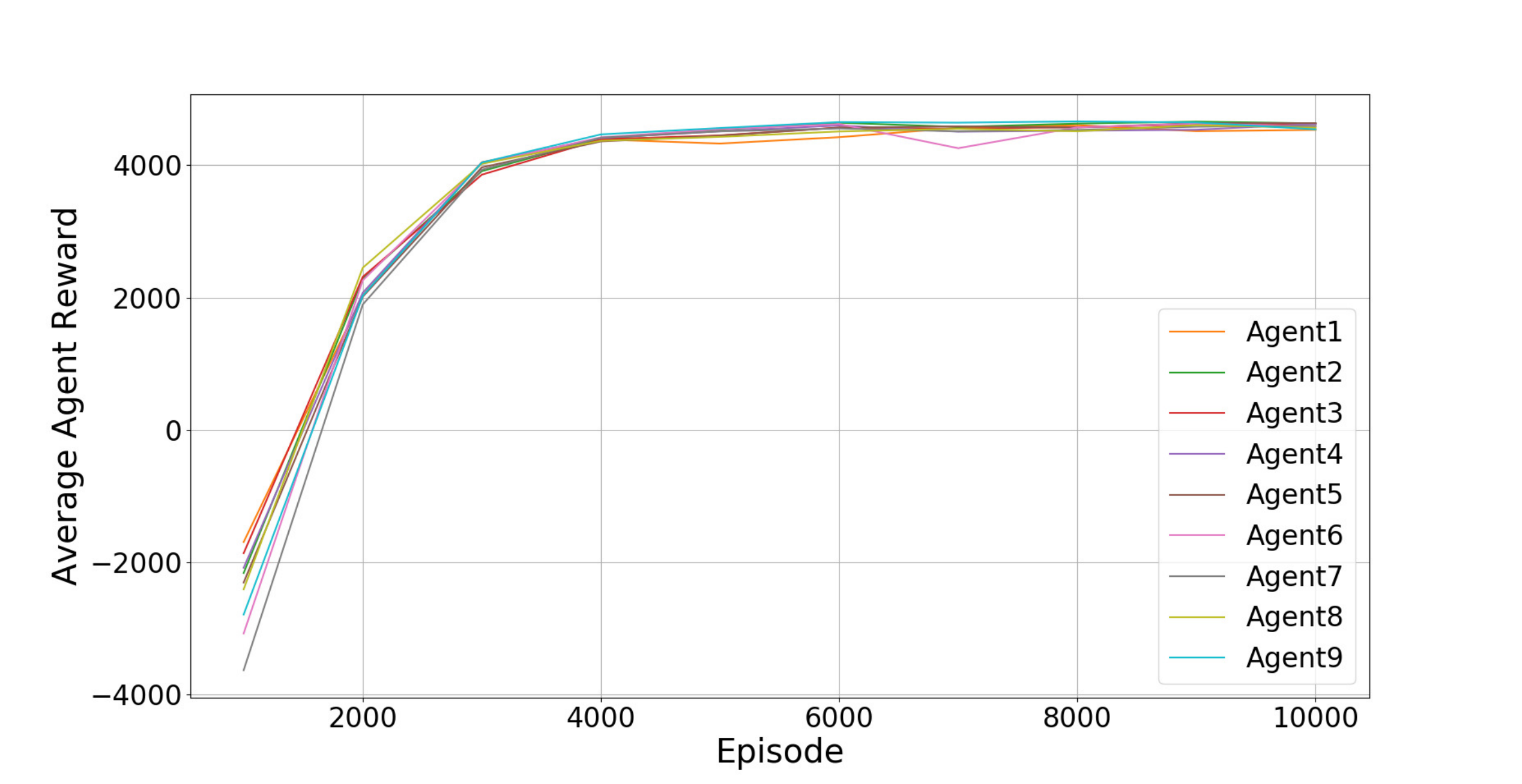}
        \caption{Average agent reward for each of the 9 agents constituting the multi-agent system}
        \label{ave_coverHorRect}
    \end{subfigure}
    \begin{subfigure}[b]{0.99\linewidth}
        \centering
        \includegraphics[width=\linewidth]{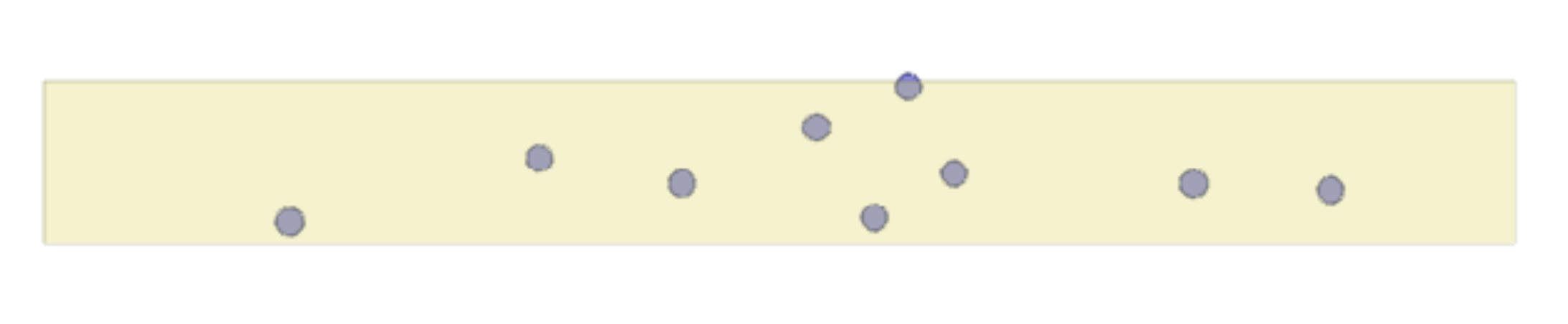}
        \caption{Qualitative performance for the area coverage}
        \label{cov_HorREct_perform}
    \end{subfigure}
\caption{Results of CR-MARL for coverage of a rectangle region.}
\label{cov_HorREct}
\end{figure}

For this set of simulations with the coverage-based reward function, since the reward function benefits from intrinsic collective concepts, the modified MADDPG algorithm is considered. The simulation results considering coverage concepts are given in Fig.~\ref{cov_sq} and Fig.~\ref{cov_HorREct} for area coverage of square and rectangle regions respectively. 

As mentioned above, in both scenarios, the basic concepts from the MADDPG algorithm~\cite{Lowe2017} are used. Exceptionally, in CR-MARL, where the modified MADDPG is applied, the original reward sharing approach is slightly amended. In contrast to the original MADDPG algorithm where in collaborative scenarios all agents use a shared collective reward, CR-MARL uses the concept of ``overall area coverage" that plays the role of reward sharing channel. In CR-MARL, the reward called for each agent is including some aspects of  individual reward along with some aspect of collective reward. On the other hand, by considering the individual aspect of reward, some levels of decentralization is automatically integrated.

As it can be observed from the simulation results, (See Fig.~\ref{cov_sq} and Fig.~\ref{cov_HorREct}), the CR-MARL scenario stands out because:
\begin{enumerate}
    \item It is showing a reasonable convergence on average rewards (see Fig.~\ref{ave_coverHorRect}) even with less number of episodes (around 5,000 episodes) compared to SW-MARL (still no convergence in 10,000 episodes). 
    \item The qualitative view of performance shows a larger spread of agents in Fig.~\ref{cov_HorREct_perform} vs. Fig.~\ref{sw_HorREct_perform}.   
\end{enumerate}
    
\subsection{Illustrative experiment}\label{illExp}
\begin{figure}[tp]
\centerline{\includegraphics[width=80mm]{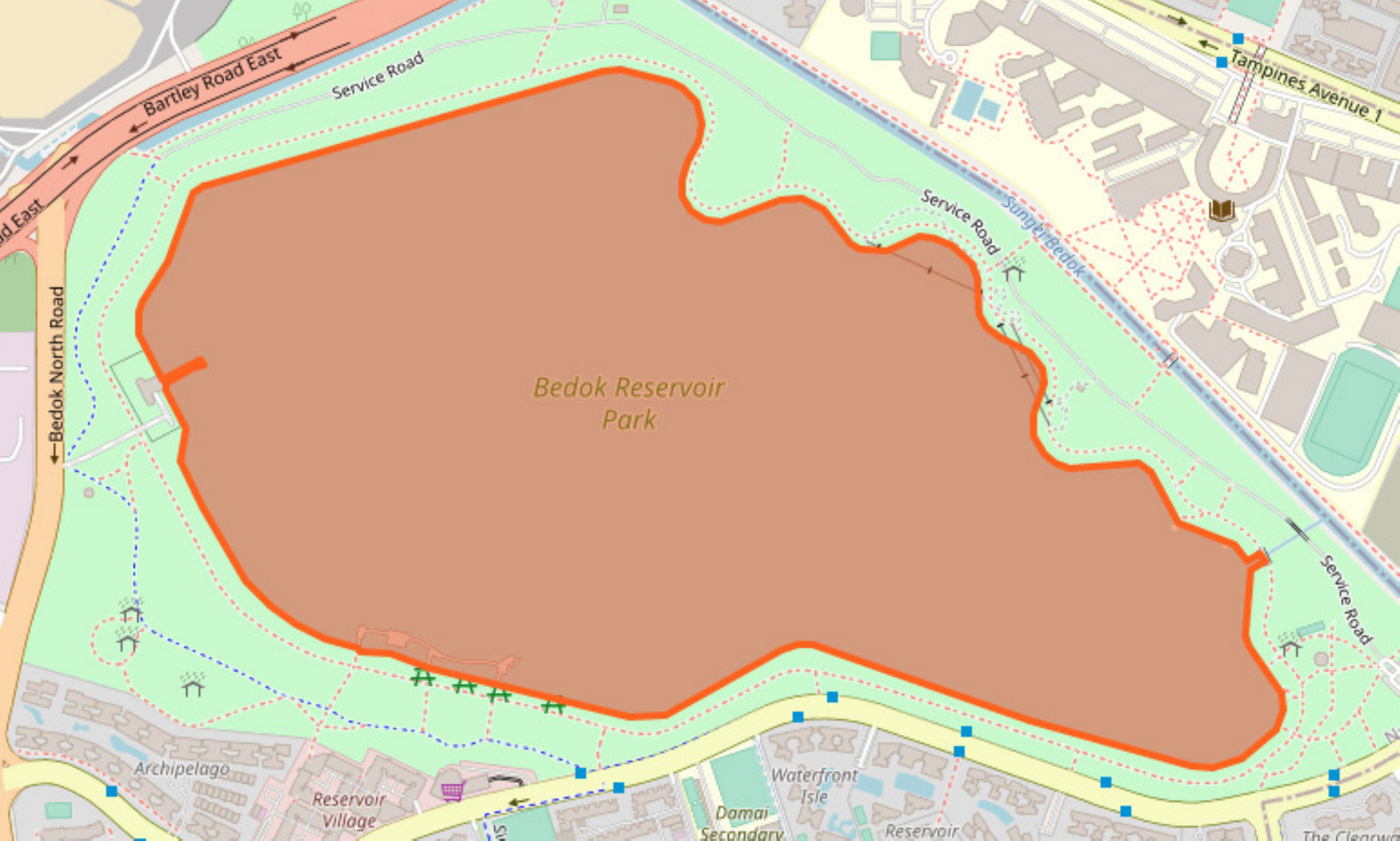}}
\caption{Bedok Reservoir map}
\label{bedok-map}
\end{figure}

\begin{figure}[htbp]
\centering
    \begin{subfigure}[b]{0.99\linewidth}
        \centering
        \includegraphics[width=\linewidth]{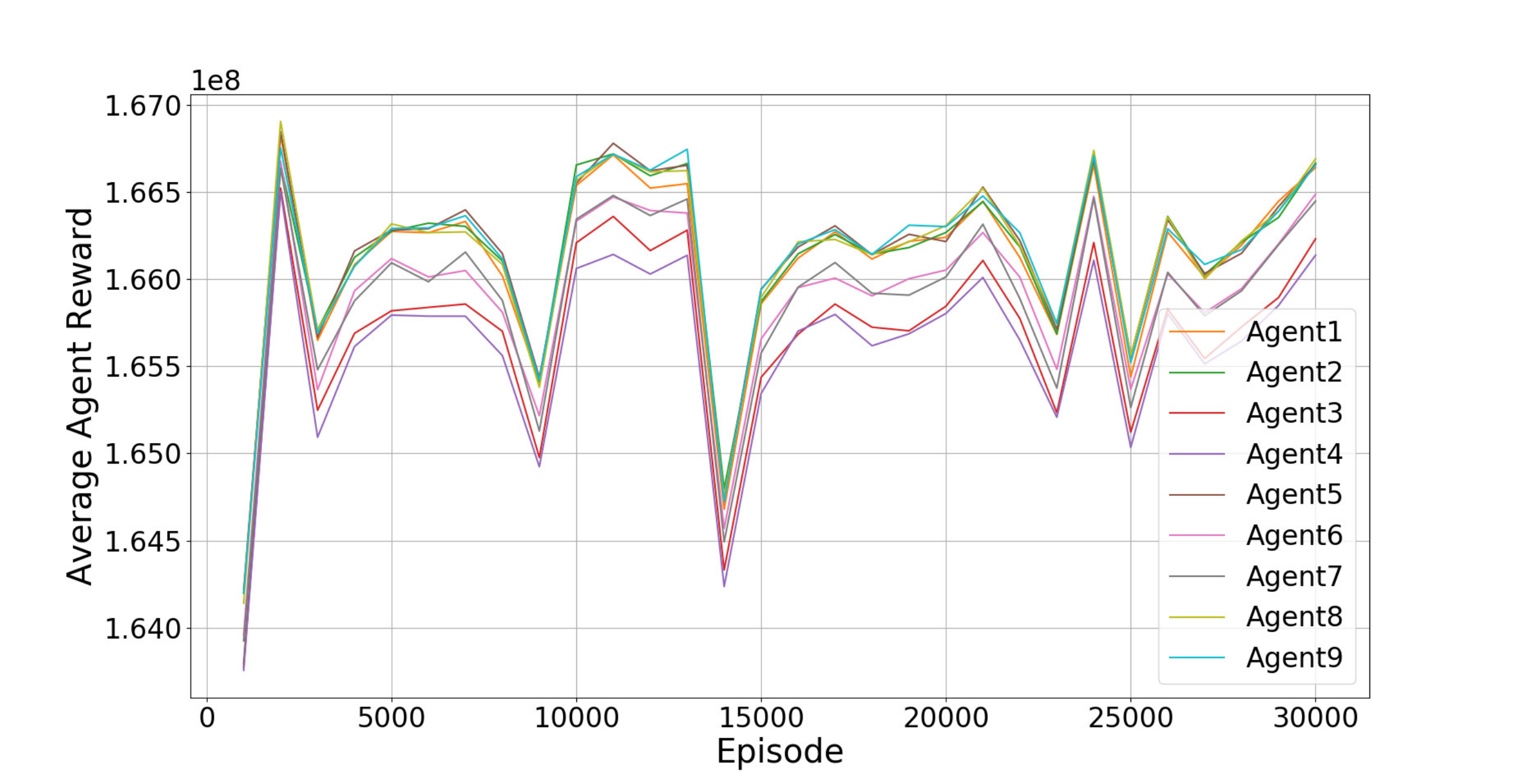}
        \caption{Average agent reward for  each  of  the  9  agents  constituting  the multi-agent system}
        \label{ave_cov_bedok}
    \end{subfigure}
    \begin{subfigure}[b]{0.99\linewidth}
      \begin{subfigure}[b]{0.99\linewidth}
        \centering
        \includegraphics[width=0.8\linewidth]{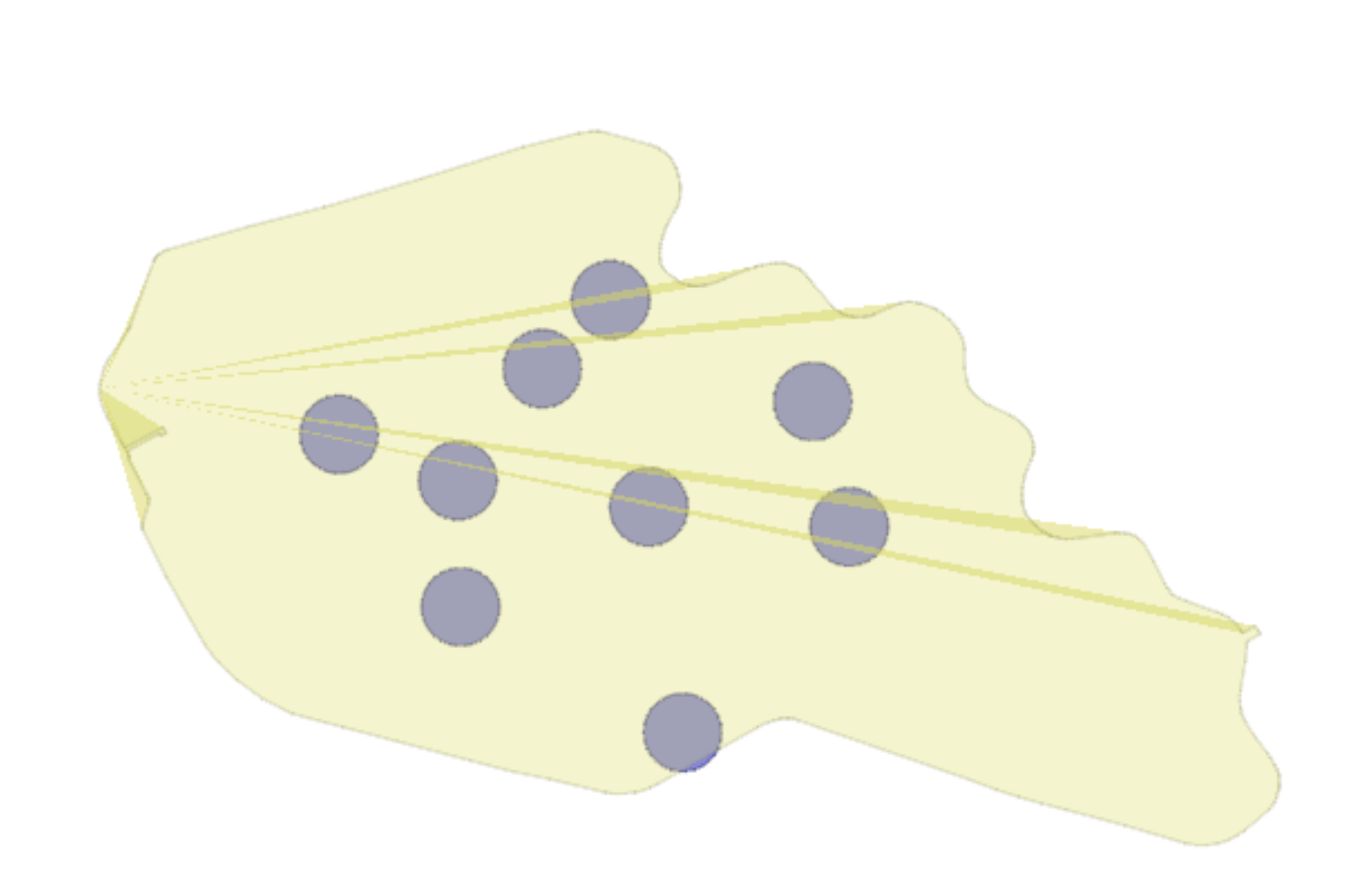}
        \label{cov_bedok_perform}
        \end{subfigure}
        \begin{subfigure}[b]{0.99\linewidth}
        \centering
        \label{cov_bedok2}
        \end{subfigure}
        \caption{Qualitative spread results of CR-MARL in Bedok Reservoir} 
        \label{cov_bedok_spread}
    \end{subfigure}
\caption{Results of CR-MARL-Coverage of Bedok Reservoir} 
\label{bedok}
\end{figure}

In this section, a real waterbody is considered as the region aimed to be covered. Bedok Reservoir is a reservoir in the eastern part of Singapore which has a surface area of 880,000~m$^2$. The map of Bedok Reservior in lon/lat coordinates is imported from OpenStreetMap database (see Fig.~\ref{bedok-map}), and after extarcting the ShapeFile through MyGeodata Converter, it is finally converted in UTM coordinates as a 2D surface. Considering bedok reservoir as the target area, the results of applying Modified MADDPG on CR-MARL are given in Fig.~\ref{bedok}.

\section{conclusion and Future Works}\label{conc}
In this paper, a collaborative framework is considered for a swarm of agents which are tasked to cover areas of varying shape. To do this, area coverage problem is modeled as a policy-based MARL approach with continuous state-space and action-space. Furthermore, two reward schemes are designed based on the required application (SW-MARL and CR-MARL), and also a modified form of the MADDPG algorithm for collaborative context is introduced for CR-MARL. The adaptation specifically targets the reward sharing policy, where the aim is intrinsically fulfilled due to the nature of the designed reward with collaborative attributes. This eliminates the need for applying a reward sharing concept as a summation of all agents' rewards. The simulation results show that both reward schemes yield an acceptable coverage performance. Owing to the systematic learning and dealing with non-stationarities in the environment, both approaches are more effective than mere swarming. However, in terms of convergence and more effective spread, CR-MARL with Modified MADDPG outperforms the SW-MARL approach.
As a scope for future work, this study can be generalized with other novel ideas incorporated with reward function engineering such as integrating neighborhood-based concepts. In addition, information sharing among agents is extendable to boundaries beyond reward sharing, and to this aim, a communication protocol among agents is required. Deciding on the topology of the communication network, and also considering the communication range in practice together with sensory/coverage range offer promising new ideas in the field of learning networked agents.



\end{document}